# Deep Learning Techniques for Super-Resolution in Video Games

Alexander Watson
*Department of Computing and Informatics*
*Bournemouth University*
Bournemouth, UK
i7600327@bournemouth.ac.uk

*Abstract*—The computational cost of video game graphics is increasing and hardware for processing graphics is struggling to keep up. This means that computer scientists need to develop creative new ways to improve the performance of graphical processing hardware. Deep learning techniques for video super-resolution can enable video games to have high quality graphics whilst offsetting much of the computational cost. These emerging technologies allow consumers to have improved performance and enjoyment from video games and have the potential to become standard within the game development industry.

## I. Introduction

Video game graphics are improving. Complex particle physics and ray-traced light effects, where environmental lighting is rendered in real time as individual rays, have become the standard within major game development studios [1]. These effects use a lot of processing power. For consumers to run games with these graphics on their computers, they require dedicated hardware called graphics processing units (GPU). Yet increases in GPU processing power are beginning to plateau as the size of transistors reaches atomic level. Moore's law, which claimed that the number of transistors on a microchip will double approximately every two years [2], is coming to an end, so computer scientists are devising new methods to improve the performance of GPUs using emerging technologies such as artificial intelligence (AI) [3].

One such AI technology is using deep neural networks to improve video game graphics in real time through a technique called video super-resolution (SR). SR is a technique where low resolution images can be upscaled to high resolution, improving the quality. This has the potential to reduce the GPU power needed to play video games, whilst simultaneously improving the graphical quality and performance. According to Wang et al [4] these methods of SR can "often achieve the state-of-the-art performance on various benchmarks".

## II. Background

Video game enthusiasts value stunning visuals in addition to high performance in their games which is a hard balance to maintain. Video resolution can impact player enjoyment, however frame rate, denoted as frames per second (fps), impacts both player performance as well as player enjoyment [5].

Playing a video game at high resolutions, such as 4k, with many graphical effects has a performance cost tied to it which can lower the frame rate to a level which the game cannot be played enjoyably. Fast paced action games require a high frame rate for a smooth gameplay experience, especially in competitive scenarios where split-second reactions can be the difference between a win or a loss. Research by Claypool et al [5] shows that in a first-person shooter game, at low frame rates such as 3 fps to 7 fps the players could not target the opponents correctly and improvements in player performance were seen up to 60 fps, at which players performed 7 times better than at the low frame rate of 3 fps.

Consumers can increase their frame rate by reducing their graphical settings or running games in a lower video resolution, however this is undesirable for the player as they must play the game with decreased visual fidelity and for the game developer as their product is not being experienced as they intended. However, deep learning video SR techniques permit a player to run the game at a lower resolution such as 1080p or 1440p, increasing their frame rate, whilst neural networks upscale the image quality to 4k, giving the desired outcome of high frame rate together with stunning visuals.

This also allows for consumers with lower specification GPUs to play games which their computers do not have the power to run at the default resolution. This makes video games more accessible for people with a lower budget, whilst also improving the performance and enjoyment for consumers with high specification GPUs.

The frontrunner in a consumer solution to this problem is NVIDIA [6] with their coined "Deep Learning Super Sampling (DLSS)" method of rendering which they claim can deliver performance increases of 2 to 3 times at 4k resolution. Alternatively, there are promising open-source solutions by companies such as AMD [7] with their "FidelityFX" technology as well as Microsoft with their up-and-coming application programming interface (API) called DirectML in which they have demonstrated "a sample that uses DirectML to execute a basic super-resolution model to upscale video from 540p to 1080p in real time" [8].

## III. Questions

There are questions to be addressed with this emerging technology which will determine it's potential to become the industry standard used by the majority of major game development studios. What are the percentage increases in frame rate using deep learning video SR? How accurately does the upscaled image compare with the original image? Are there any other performance costs? What hardware is required to make use of these solutions at a consumer level and furthermore how accessible are these solutions for the game developers?

To answer these questions, we will look at research on deep learning for image super-resolution by Wang et al [4],

focusing on video super-resolution with convolutional neural networks [9], how motion compensation can be used [10] and the computational cost of these techniques [11] in real time [12]. We will also look at the current solutions from NVIDIA [6], [13], [14], AMD [7] and Microsoft [8].

## IV. FRAME RATE

The most substantial source of data on frame rate increase using deep learning video SR is NVIDIA's official Deep Learning Super-Sampling (DLSS) information page, where they display figures showing the average frames per second (fps) with and without DLSS on four different video games at three different resolutions across their entire "RTX 2000" series of graphics processing units (GPU).

For a game called 'Control', there are large increases in frame rate for every resolution on each of the tested GPUs. At 1080p the fps increases by a range of 51-63% across the various GPUs, with the smallest increase coming from the then flagship RTX 2080 Ti GPU with an average of 77.8 fps with DLSS off and 117.4 fps with DLSS on and the largest increase coming from the cheapest RTX 2060 GPU going from an average of 41.8 fps with DLSS off to 68.1 fps with DLSS on. At 4k the frame rate increase is drastically higher with an increase ranging from 159-360% across the GPUs, with the RTX 2080 Ti increasing from an average of 25.8 fps to 69.2 fps and the RTX 2060 improving from 8 fps to 36.8 fps with 360% total performance increase as seen in Fig 1.

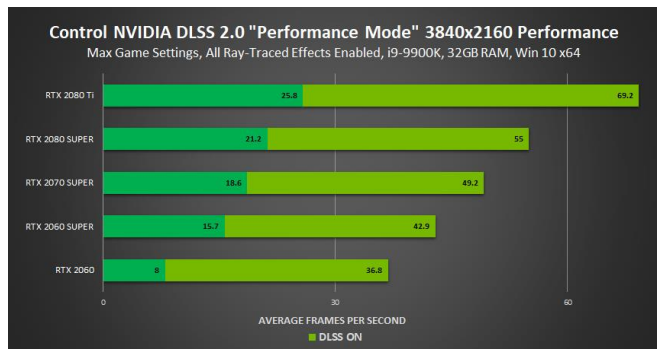

Fig. 1. Graph showing FPS increases in 'Control' at 3840x2160 (4k) resolution with DLSS off and DLSS on across several NVIDIA GPUs [13].

This trend continues across the three other games tested with 'Deliver Us the Moon' having increases ranging from 30-56% at 1080p and 126-157% at 4k, 'MechWarrior 5: Mercenaries' having an increase of 27-37% at 1080p and 103-135% at 4k and 'Wolfenstein: Youngblood' having an increase of 22-34% at 1080p and 82-97% at 4k.

From this data we can see that the fps increases are substantial, with generally greater percentage increases coming from the games with lower initial average fps with DLSS turned off. The percentage increases at higher initial fps were smaller but still substantial. These results show that deep learning SR can be very effective in improving the performance of video games, with NVIDIA's DLSS even able to make their cheapest RTX 2060 GPU have greater performance with DLSS turned on than their flagship RTX 2080 Ti GPU does with DLSS turned off in some examples, despite these GPUs having a difference in price of 3 to 4 times.

## V. ACCURACY

### A. Convolutional neural networks

Convolutional neural networks (CNNs) are an efficient technique for video super-resolution with deep learning. CNNs require less pre-processing than traditional image SR techniques and can be trained to identify and differentiate between key features in each frame [10]. For video SR, CNNs can be given the adjacent frames as input which assists the neural network to learn the spatial and temporal information of the frame, which improves the CNNs ability to produce high quality results [9].

### B. Motion compensation

Motion compensation is critical when using CNNs for video SR. When there is a large amount of motion between two adjacent frames it can be difficult for the neural network to produce a smooth video and can lead to motion blur in the result [9]. Providing the CNN with motion vectors allows for the algorithm to compensate for the motion between frames and produce a high-quality SR video [10].

### C. Deep Learning Super Sampling

NVIDIA's method for video SR takes low-resolution aliased frame images as well as their motion vectors, which have information about the direction of movement of every pixel in the following frame, and inputs them into a CNN called a convolutional autoencoder which is trained on a supercomputer to use these inputs to produce a high-resolution version of the frame image on a frame-by-frame basis [6].

Video produced using SR techniques will never be as accurate as a high-resolution original as machine learning techniques are based on approximation. The accuracy of SR is tested during neural network training through the use of algorithms called 'loss functions' which compare the image produced by the neural network to a high-resolution native image called a ground-truth image [4], measuring against different attributes of the image and then using the results to further train the neural network to produce a more accurate result. For NVIDIA's DLSS "this process is repeated tens of thousands of times on the supercomputer until the network reliably outputs high quality, high resolution images" [6].

### D. Image Quality Assessment

The results of deep learning SR can be evaluated using performance metrics called image quality assessments (IQA). There are two main types of IQA which vary greatly in their assessment results. There are subjective methods which can be used such as a human simply comparing the video visually and evaluating the quality of the produced images. These methods are regarded as the most useful as human perception directly relates to how the player would perceive the image they see when playing the game. However, this method of IQA is very time consuming so the alternative method is more commonly used [4]. Objective IQA methods are using computational models to compare the images. The problem with these methods is that the computer does not perceive in the same way as humans, so the assessment results can easily contradict the results of subjective IQA [4]. This makes the task of performance measurement difficult to compute effectively, however video

SR using deep learning can produce effective results based upon both subjective and objective IQA methods [12].

## VI. Performance Costs

An advantage of video SR over single-image SR is that the adjacent frames can be used alongside the current frame and motion vectors to reduce the computation needed to generate the current frame. Adaptive transfer can be used to help the convolutional neural network (CNN) learn how to process the motion information in the motion vectors from previous frames. This greatly increases the speed of the algorithm with negligible loss of performance [11].

For NVIDIA's DLSS, the CNN is trained by their supercomputer ahead of time and delivered to the consumer's GPU through online software in driver updates, so most of the computation is carried out before the video game is played and there are only performance gains to be had by using their solution in real time [6].

## VII. Accessibility

### A. Consumers

The only solution which is widely available to the consumer at present is NVIDIA's DLSS which requires specific dedicated hardware in the form of one of their RTX series graphics processing units (GPU). As of the time of writing these range from £281.99 for the entry level RTX 2060 to £1,399.00 for the new flagship RTX 3090 when bought from NVIDIA's online store directly [14]. This price is for the GPU only which is just a single component of a personal computer (PC) making the total price for entry much greater. This makes it so that currently the DLSS solution is only available to enthusiast PC owners with mid-range to high-range PCs. However, as this hardware becomes superseded by new hardware, the prices will reduce making it more accessible to consumers with lower budgets.

Other solutions such as AMD's and Microsoft's are still in development and it is unknown what hardware these will require. However, an early version of Microsoft's DirectML has already been made available for anyone to use as an open-source API on GitHub, with no existing hardware requirements as the super-resolution is performed using only software [8].

### B. Developers

For deep learning video SR to become a standard technique used in game development, these techniques need to be widely available for use by game developers globally. According to NVIDIA, "DLSS 2.0 is now available to Unreal Engine 4 developers through the DLSS Developer Program that will accelerate deployment in one of the world's most popular game engines" [6]. Unreal Engine 4 is one of the most popular game engines [15], making these tools widely available internationally. Additionally, AMD's FidelityFX [7] and Microsoft's DirectML [8] are both open-source tools available for anyone to use although they are in their infancy compared to NVIDIA's solution.

## VIII. Conclusion

There is potential for deep learning super-resolution techniques to become standard in the video games industry, allowing consumers to have substantial increases in their GPUs performance whilst maintaining high quality visuals. These emerging technologies are still in their early stages and currently there is only support available for NVIDIA's Deep Learning Super Sampling (DLSS) for a limited number of games. However, the number of supported titles is increasing steadily in addition to open-source tools from Microsoft and AMD being made available. The frame rate increases are substantial, especially in the cases where the frame rate is very low without these techniques being utilised, which is where a greater frame rate is needed the most. With increases in player performance up to 60 fps, consumers can benefit from DLSS helping them get closer to or above this threshold. Adjacent frames and motion vectors being available as input to the CNN allows for greater accuracy of the resulting video, with good results based upon subjective IQA methods. The performance cost of the deep learning algorithms is negligible, with the performance gains outweighing these greatly. The barrier to entry is quite high, with the suitable GPUs costing a minimum of £281.99, but the supported graphics processing units (GPU) will become cheaper as this hardware is superseded.

Many of the questions raised in this paper have been answered to an extent, but due to the topic being a state-of-the-art emerging technology, there is limited data available to manifest a full conclusion. The only available frame-rate statistics testing deep learning SR for video games are provided by NVIDIA, which have likely given best case results to promote their product which may not be entirely replicable by a consumer. Testing the accuracy of the solutions is also difficult to accomplish due to the best metric to evaluate results being subjective and based upon human perception.

## IX. Future Work

These questions could be answered further with future research which could be conducted by independently testing NVIDIA's DLSS solution against AMD and Microsoft's solutions in a fair environment where the variables can be controlled. The accuracy of the results could be evaluated in a single-blind study by a large number of human subjects which could give a fair analysis on each of the video super-resolution methods based on their perception without any bias.